\documentclass[lettersize,journal]{IEEEtran}
\usepackage{amsmath,amsfonts}
\usepackage{algorithmic}
\usepackage{algorithm}
\usepackage{array}
\usepackage[caption=false,font=normalsize,labelfont=sf,textfont=sf]{subfig}
\usepackage{textcomp}
\usepackage{stfloats}
\usepackage{url}
\usepackage{verbatim}
\usepackage{graphicx}
\usepackage{cite}
\usepackage{booktabs}
\begin{document}

\title{AMMUNet: Multi-Scale Attention Map Merging for Remote Sensing Image Segmentation}

\author{Yang Yang, Shunyi Zheng
\thanks{This work was supported by the National Key Research and Development Program of China under Grant No.2022YFF0904400.}
\thanks{Yang Yang and Shunyi Zheng are with the School of Remote Sensing and Information Engineering, Wuhan University, Wuhan 430079, China (e-mail: yangyang001@whu.edu.cn).}}

\markboth{Journal of \LaTeX\ Class Files,~Vol.~14, No.~8, August~2021}%
{Shell \MakeLowercase{\textit{et al.}}: A Sample Article Using IEEEtran.cls for IEEE Journals}

\IEEEpubid{0000--0000/00\$00.00~}

\maketitle

\begin{abstract}
The advancement of deep learning has driven notable progress in remote sensing semantic segmentation. Attention mechanisms, while enabling global modeling and utilizing contextual information, face challenges of high computational costs and require window-based operations that weaken capturing long-range dependencies, hindering their effectiveness for remote sensing image processing. In this letter, we propose AMMUNet, a UNet-based framework that employs multi-scale attention map merging, comprising two key innovations: the granular multi-head self-attention (GMSA) module and the attention map merging mechanism (AMMM). GMSA efficiently acquires global information while substantially mitigating computational costs in contrast to global multi-head self-attention mechanism. This is accomplished through the strategic utilization of dimension correspondence to align granularity and the reduction of relative position bias parameters, thereby optimizing computational efficiency. The proposed AMMM effectively combines multi-scale attention maps into a unified representation using a fixed mask template, enabling the modeling of global attention mechanism. Experimental evaluations highlight the superior performance of our approach, achieving remarkable mean intersection over union (mIoU) scores of 75.48\% on the challenging Vaihingen dataset and an exceptional 77.90\% on the Potsdam dataset, demonstrating the superiority of our method in precise remote sensing semantic segmentation. Codes are available at https://github.com/interpretty/AMMUNet.
\end{abstract}

\begin{IEEEkeywords}
Attention map merging, global attention mechanism, remote sensing, semantic segmentation.
\end{IEEEkeywords}

\section{Introduction}
\IEEEPARstart{S}{emantic} segmentation of remote sensing imagery holds paramount importance, as it lays the groundwork for a myriad of applications spanning urban planning, environmental monitoring, and resource management. The advent of deep learning, particularly Convolutional Neural Networks (CNNs), has ushered in a remarkable paradigm shift, revolutionizing the field of computer vision and, by extension, remote sensing image analysis.

Long et al.\cite{long2015fully} pioneered the Fully Convolutional Network, establishing it as the first end-to-end trainable semantic segmentation architecture. Since then, ResNet\cite{he2016deep} revolutionized deep neural networks by introducing residual connections, allowing for effective training of very deep architectures while mitigating the degradation problem and preserving rich multi-level representations. Building upon these advances, DeepLabv3\cite{chen2017rethinking} and its enhanced version DeepLabv3+\cite{chen2018encoder} employed atrous/dilated convolutions to expand the receptive field and capture multi-scale information. However, despite these enhancements, conventional CNN-based methods still struggle to model global contextual information due to their inherently local receptive fields.

To enhance global information utilization, ViT \cite{dosovitskiy2020image} applied multi-head self-attention (MSA)\cite{vaswani2017attention} to images using window splitting and Swin Transformer \cite{liu2021swin} was proposed to exchang window information via shifted windows.
For the task of dense semantic segmentation of large-scale remote sensing imagery\cite{li2021abcnet},\cite{li2022multiattention}, Transformers and CNNs have been synergistically employed in conjunction with the U-Net\cite{ronneberger2015u} architecture, with the aim of effectively capturing both fine-grained details and global contextual information\cite{cao2022swin},\cite{he2022swin}. Efforts include MAResUNet \cite{li2021multistage} for efficiency via linear attention, and UNeFormer \cite{wang2022unetformer} combining global and local feature information through the incorporation of convolutional and attention modules in an efficient framework.
\IEEEpubidadjcol

However, despite their successes, Transformer-based approaches suffer from a significant limitation: the necessity for window-based operations to mitigate the computational demands of global self-attention mechanisms\cite{guo_attention_2022},\cite{su2023global}. 
Firstly, it weakens the contextual relationships between patches, potentially hindering the accurate modeling of long-range dependencies\cite{wang_non-local_2018}. Secondly, the relative positional relationships between windows become implicit, further complicating the modeling process. Even the shifted window approach employed by the Swin Transformer can only partially alleviate this issue by modeling relationships between adjacent windows.

\begin{figure*}[!t]
	\centering
	\includegraphics[width=\textwidth]{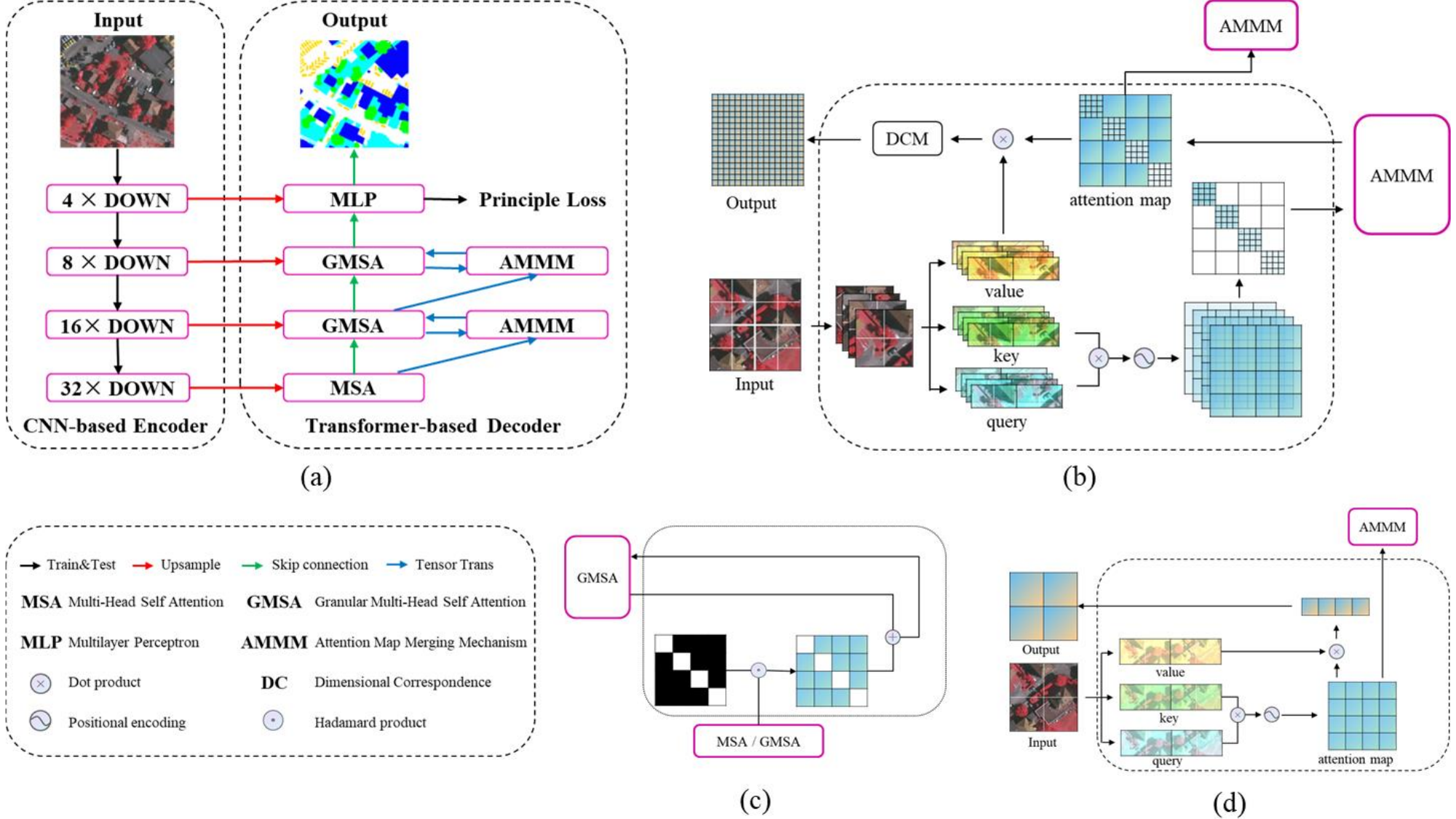}
	\caption{(a) Overview of AMMUnet. (b) Detailed structure of GMSA. (c) Detailed structure of AMMM. (d) Detailed structure of MSA.}
	\label{fig_sim}
\end{figure*}

To unlock the full potential of Transformer-based architectures for remote sensing image segmentation, we introduce AMMUNet, a UNet-based framework that employs a ResNet encoder to extract multi-scale features and an innovative attention mechanism in the decoder. This mechanism leverages the strengths of both CNNs and Transformers, reusing high-level attention maps to preserve long-range modeling capabilities while utilizing our granular multi-head self-attention (GMSA) module to compute fine-grained, local attentional relationships. Furthermore, we propose an attention map merging mechanism (AMMM) that employs a fixed mask template to merge attention maps generated at different scales, enabling the modeling of global attention mechanism while minimizing computational demands and ensuring optimal accuracy. The key innovations of our work can be summarized as follows:
\begin{enumerate}
	\item{The proposed AMMUNet is a semantic segmentation network that leverages a CNN-based ResNet encoder and a novel global attention mechanism in the decoder, capable of merging multi-scale attention maps.}
	\item{We design the GMSA module to extract attention maps at the current scale and employ the AMMM to merge multi-scale attention maps. Our approach includes strategies for granularity alignment, relative position bias reduction, and a fixed mask template for attention map merging, effectively reducing computational demands compared to global multi-head self-attention while achieving real-time performance.}
	\item{Experimental evaluations on the Vaihingen and Potsdam datasets demonstrate the superiority of our method in terms of accuracy and efficiency.}
\end{enumerate}

\section{METHODOLOGY}
\subsection{Overview of AMMUNet}
The proposed semantic segmentation network is illustrated in Fig. \ref{fig_sim}(a). The network adopts UNet as the framework, which is an overall encoder-decoder structure. The encoder employs ResNet50 as backbone to extract multi-scale features. ResNet50 is composed of four stages, each performing 2$\times$ spatial downsampling. The four outputs correspond to feature maps of different sizes, from shallow to deep levels: $\frac{H}{4} \times \frac{W}{4}$, $\frac{H}{8} \times \frac{W}{8}$, $\frac{H}{16} \times \frac{W}{16}$, and $\frac{H}{32} \times \frac{W}{32}$, respectively.

The decoder takes the deepest level feature map from ResNet as its initial input. This input is first passed through a MSA module, generating the corresponding attention map and feature map. The attention map serves as the input to our proposed AMMM module, while the feature map is upsampled and concatenated with the intermediate feature maps from ResNet via skip connections, jointly feeding into the GMSA module. In the middle two layers of the decoder, we employ the GMSA module to compute and transform the input feature maps, and the resulting attention maps are passed to the AMMM module. The AMMM module merges the attention maps from different scales to form a unified global attention map, which is then fed back into the GMSA to guide the subsequent feature map computation. Finally, the high-resolution feature map outputted by the decoder is passed through multi-layer perceptron (MLP) to perform pixel-wise semantic prediction. We calculate the loss function based on the discrepancy between the predictions and ground truth, and use backpropagation to train the entire network in an end-to-end manner.

\subsection{Global Multi-head Self-Attention}
Conventional Transformer approaches in computer vision often employ patch embeddings to divide the input image into smaller windows prior to computing self-attention, a strategy aimed at reducing computational demands. However, this comes at the cost of sparse inter-patch associations. In contrast, our approach adopts a global methodology for computing self-attention on the deepest level feature map within the decoder, as shown in Fig. \ref{fig_sim}(d). This decision is grounded in two key considerations: first, the window size at the deepest decoder level is typically modest, limiting the computational savings afforded by window-based operations; second, a global approach enables fine-grained computation of relevance between individual pixel regions.

The attention map (AM) and the attention output (Attn) are computed as follows:
\begin{gather}
\mathrm{AM}(Q,K)=\mathrm{softmax}\left(\frac{QK^T}{\sqrt{d_k}}+\mathrm{RPB}\right) \\
\mathrm{Attn} = \mathrm{AM}(Q, K)V\label{eq_Attn}
\end{gather}
where Q, K, and V represent the queries, keys, and values respectively, d denotes the input dimension, and RPB signifies the relative position bias. Notably, the attention map not only contributes to the attention output computation but also serves as an input to AMMM module, facilitating the fusion of attention maps. This approach enables the modeling of long-range dependencies between distant pixels within shallow-level attention maps. 

\subsection{Granular Multi-head Self-Attention}
As illustrated in Fig. \ref{fig_sim}(b), the GMSA module, in contrast to standard multi-head self-attention modules, divides the overall feature map into granular windows of 2$\times$2 pixels. For each subregion, it computes the queries (Q), keys (K), and values (V) to obtain the corresponding attention map, denoted as $\mathrm{AM}(Q_{2\times2}, K_{2\times2})$. Assuming an input feature map of size ${H} \times {W}$, there are $N = \frac{H}{2} \times \frac{W}{2}$subregions, yielding ${AM}_1,\ldots,{AM}_N$ individual subregion attention maps.

The relative position bias extends the concept of self-attention to encode the distance between any two tokens. Since the relative positions within each subregion are fixed, the same relative position bias can be applied across different subregions for computational efficiency.

These individual subregion attention maps are then inserted into the diagonal region of the overall attention map:
\begin{equation}
AM=\left(\begin{matrix}{AM}_1&&\\&\ddots&\\&&{AM}_N\\\end{matrix}\right)
\end{equation}
The GMSA computes attention within each subregion of the current layer, contributing to both the current layer's attention computation and representing the relevance at the corresponding granularity for shallower layers. The computation cost of Global MSA module, window MSA\cite{liu2021swin} and GMSA are:
\begin{flalign}
&\Omega(MSA)\hspace{-0.1cm}=\hspace{-0.1cm}4hwC^2 + 2(hw)^2C \\
&\Omega(W\text{-}MSA)\hspace{-0.1cm}=\hspace{-0.1cm}4hwC^2 + 2M^2hwC \\
&\Omega(GMSA)\hspace{-0.1cm}=\hspace{-0.1cm}4hwC^2 + (h_0w_0)^2C + 16\log_2{(\frac{hw}{h_0w_0})}C
\end{flalign}
where hw denotes the spatial dimensions of the input feature map, $M$ be the window size, and $h_0w_0$ represent the dimensions of the deepest level feature map in the decoder. In contrast to the MSA module and W-MSA, the proposed GMSA exhibits a slower computational complexity growth rate as the input scale increases, making it particularly well-suited for large-sized remote sensing imagery feature extraction.

\subsection{Attention Map Merging Mechanism}
The AMMM is designed to merge the attention maps computed by GMSA at the current scale with those propagated from deeper layers, either from a patch-based MSA or a preceding GMSA module, as shown in Fig. \ref{fig_sim}(c). The deeper-level attention maps are first multiplied by a fixed mask template, reserving space for the current-scale attention map, and then upsampled, as expressed by the following equation:
\begin{equation}
{AM}_i=\left(1-E\right){AM}_{i-1}+E\circ\ AM
\end{equation}
where $E$ denotes the fixed mask template implemented as an identity matrix, $\circ$ represents the operator of Hadamard product, $AM$ represents the current-scale attention map computed by GMSA, and ${AM}_{i-1}$ signifies the attention map from the previous module, representing the merged deeper-level attention map.

\subsection{Dimension Correspondence Module}

The Dimension Correspondence Module (DCM) facilitates the conversion of attention map storage in memory from a window-based format to a row-based format. For shallow-level features, attention maps are stored in a window-based manner. However, as AMMM merges attention maps across multiple scales, and the attention output computed via Equation \ref{eq_Attn} is stored in a row-based format, a scale alignment is necessary. This is achieved through the following dimension transformation:
\begin{equation}
{Attn}_{i\times j,k\times l}=reshape\_permute({Attn}_{i,k,j\times l})
\end{equation}
where $i,j,k,l$ represent the dimensions of Attn', with subregion sizes of 2$\times$2, corresponding to dimension sizes of $[B, \frac{H}{2}, \frac{W}{2}, 2, 2]$. Multiple DCM operations may be required, consistent with the number of upsampling layers in the architecture.

\begin{table*}[htbp]
	\centering
	\caption{EXPERIMENTAL RESULTS OF THE VAIHINGEN DATASET}
	\resizebox{\linewidth}{!}{
		\begin{tabular}{lccccccccc}
			\toprule
			Methods & Backbone & Imp.surf & Building & Low veg. & Tree & Car & Clutter & mIoU & mAcc \\
			\midrule
			DeepLabV3+ & R50 & 86.03 & 91.52 & 70.91 & 79.71 & 75.11 & 42.87 & 74.36 & 81.23 \\
			DANet & R50 & 85.12 & 91.03 & 70.72 & 79.78 & \textbf{75.65} & 38.87 & 73.53 & 80.80 \\
			OCRNet & HR18s & 85.07 & 90.38 & 70.48 & 79.77 & 72.23 & 41.03 & 73.16 & 80.46 \\
			SegFormer & Mit-B1 & 86.04 & 91.81 & \textbf{72.34} & \textbf{80.44} & 75.38 & 31.52 & 72.92 & 80.22 \\
			UNetFormer & R50 & 85.30 & 91.20 & 71.18 & 79.89 & 74.88 & 46.61 & 74.84 & 82.39 \\
			Ours & R50 & \textbf{86.14} & \textbf{91.83} & 71.21 & 79.76 & 73.98 & \textbf{49.98} & \textbf{75.48} & \textbf{82.97} \\
			\bottomrule
		\end{tabular}
	}
	\label{tab:vaihingen_results}
\end{table*}
\begin{table*}[htbp]
	\centering
	\caption{EXPERIMENTAL RESULTS OF THE POTSDAM DATASET}
	\resizebox{\linewidth}{!}{
		\begin{tabular}{lccccccccc}
			\toprule
			Methods & Backbone & Imp.surf & Building & Low veg. & Tree & Car & Clutter & mIoU & mAcc \\
			\midrule
			DeepLabV3+ & R50 & 85.73 & 92.22 & 76.10 & 77.90 & \textbf{90.72} & 41.56 & 77.37 & 85.10 \\
			DANet & R50 & 86.08 & 91.84 & 74.72 & 77.74 & 88.42 & 41.51 & 76.72 & 85.14 \\
			OCRNet & HR18s & 85.37 & 90.82 & 74.48 & 76.60 & 88.81 & 40.09 & 76.03 & 84.04 \\
			SegFormer & Mit-B1 & 85.80 & 91.90 & \textbf{76.48} & 77.72 & 88.83 & 41.46 & 77.03 & 84.77 \\
			UNetFormer & R50 & \textbf{86.46} & 92.27 & 75.14 & 78.21 & 90.10 & 42.11 & 77.38 & 85.04 \\
			Ours & R50 & 86.38 & \textbf{92.40} & 75.62 & \textbf{78.77} & 89.92 & \textbf{44.34} & \textbf{77.90} & \textbf{85.92} \\
			\bottomrule
		\end{tabular}
	}
	\label{tab:potsdam_results}
\end{table*}

\section{EXPERIMENT}
\subsection{Dataset}
The segmentation model was evaluated using two datasets: the Vaihingen dataset and the Potsdam dataset. The Vaihingen dataset comprises 33 true orthophoto images, with 17 for training and the rest for testing. Image dimensions vary between 800 and 3000 pixels. The Potsdam dataset consists of 38 images, each sized at 6000×6000 pixels. We use 24 images for training and 14 for testing. Both datasets have a ground sampling distance of 9 cm. Annotations include pixel-level labels for six categories: Impervious Surface, Buildings, Low Vegetation, Trees, Cars, and Clutter. To facilitate GPU training, we cropped the images and labels into non-overlapping 512×512 pixels.

\subsection{Evaluation Metrics}
To evaluate the performance of our proposed AMMUNet framework, we employ two widely adopted metrics: mean intersection-over-union (mIoU) and mean pixel accuracy (mAcc)\citen{guo_skysense_2023}. These metrics range from 0 to 1, with higher values indicating better model performance. Notably, mIoU is a more stringent metric for imbalanced datasets and is more commonly used for semantic segmentation tasks.
The mIoU is calculated as follows:
\begin{equation}
\mathrm{IoU} = \frac{\mathrm{TP}}{\mathrm{TP} + \mathrm{FP} + \mathrm{FN}}
\end{equation}
\vspace{-0.3cm}
\begin{equation}
\mathrm{mIoU} = \frac{1}{C} \sum_{i=1}^{C} \mathrm{IoU}_i
\end{equation}
The mAcc is defined as:
\begin{equation}
\mathrm{Acc} = \frac{\mathrm{TP} + \mathrm{TN}}{\mathrm{TP} + \mathrm{FP} + \mathrm{FN} + \mathrm{TN}}
\end{equation}
\vspace{-0.3cm}
\begin{equation}
\mathrm{mAcc} = \frac{1}{C} \sum_{i=1}^{C} \mathrm{Acc}_i
\end{equation}
where TP, FP, FN, and TN represent the true positive, false positive, false negative, and true negative pixel counts, respectively, and C denotes the total number of classes in the dataset.

\subsection{Implementation Details}

All experiments were conducted using the PyTorch framework on a single NVIDIA GTX 3090 GPU. During the training phase, we utilized the AdamW optimizer with a weight decay parameter of 0.01. The maximum training epoch count was set to 200, with a training batch size of 8. The initial learning rate was 1e-4, and the loss function employed was cross-entropy. We updated the learning rate using the "poly" learning rate strategy, where the power is 0.9.

\subsection{Comparison with State-of-the-Art Methods}

To validate the superiority of our proposed model, we benchmarked it against five state-of-the-art methods, including DeeplabV3+\cite{chen2018encoder}, DANet\cite{fu2019dual}, OCRNet\cite{yuan2020object}, SegFormer\cite{xie2021segformer}, and UNetFormer\cite{wang2022unetformer}. DeeplabV3+, a purely CNN-based architecture, is widely regarded as the baseline for semantic segmentation and enhances the model's receptive field using dilated convolutions. The remaining methods combine CNN and Transformer architectures, leveraging self-attention mechanisms to capture global semantic contexts. Leveraging the advantages on remote sensing segmentation tasks, our approach incorporates UNetFormer as a baseline and conducts experiments to evaluate the performance of our proposed method against this baseline network.
To ensure a fair evaluation, all networks employed ResNet50 as the backbone, except for OCRNet and SegFormer, which utilized HRNet18s and MixTransformerB1, respectively, due to their similar model sizes to ResNet50.
We evaluated these models using mIoU and mAcc as our evaluation metrics on the Vaihingen and Potsdam datasets. Detailed results, including category-specific IoU metrics, are presented in Tables \ref{tab:vaihingen_results} and \ref{tab:potsdam_results}.

\begin{figure}[t]
	\centering
	\includegraphics[width=\linewidth]{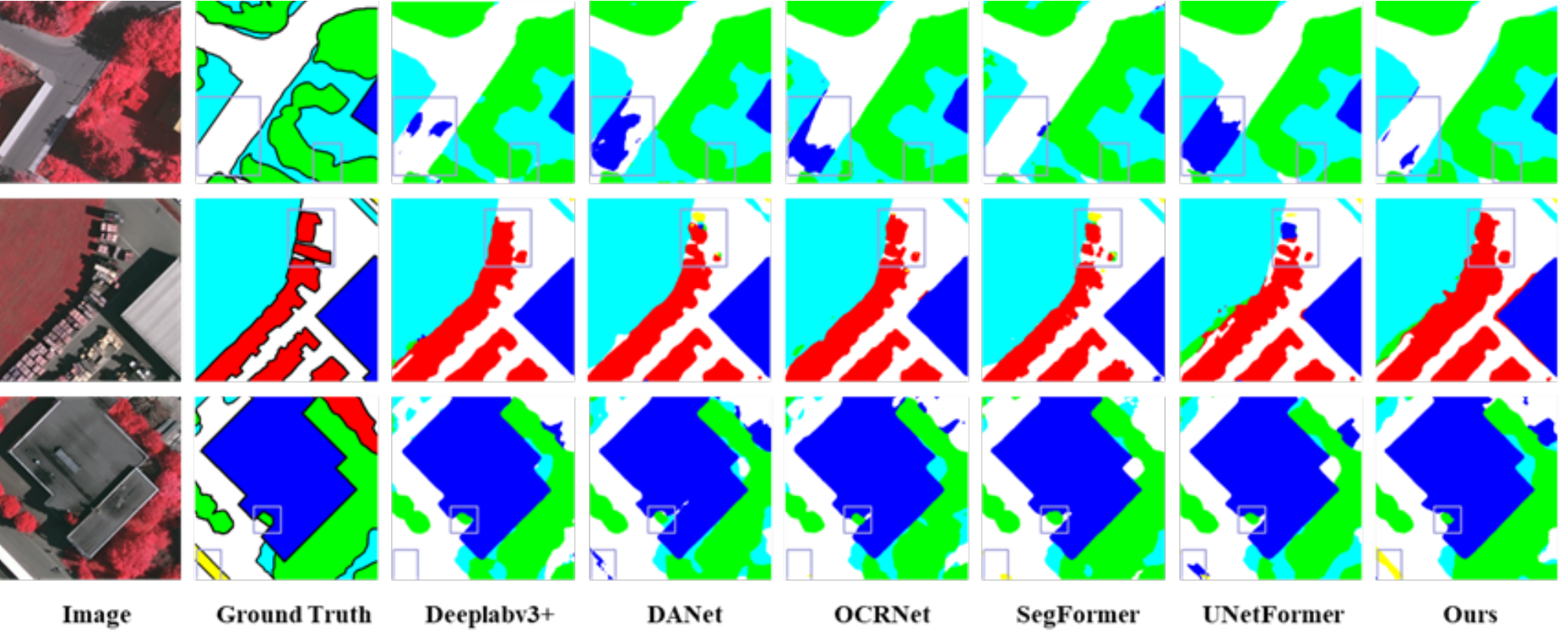}
	\caption{Visualization results on the vaihingen dataset.}
	\label{fig_vis_vai}
\end{figure}

\begin{figure}[t]
	\centering
	\includegraphics[width=\linewidth]{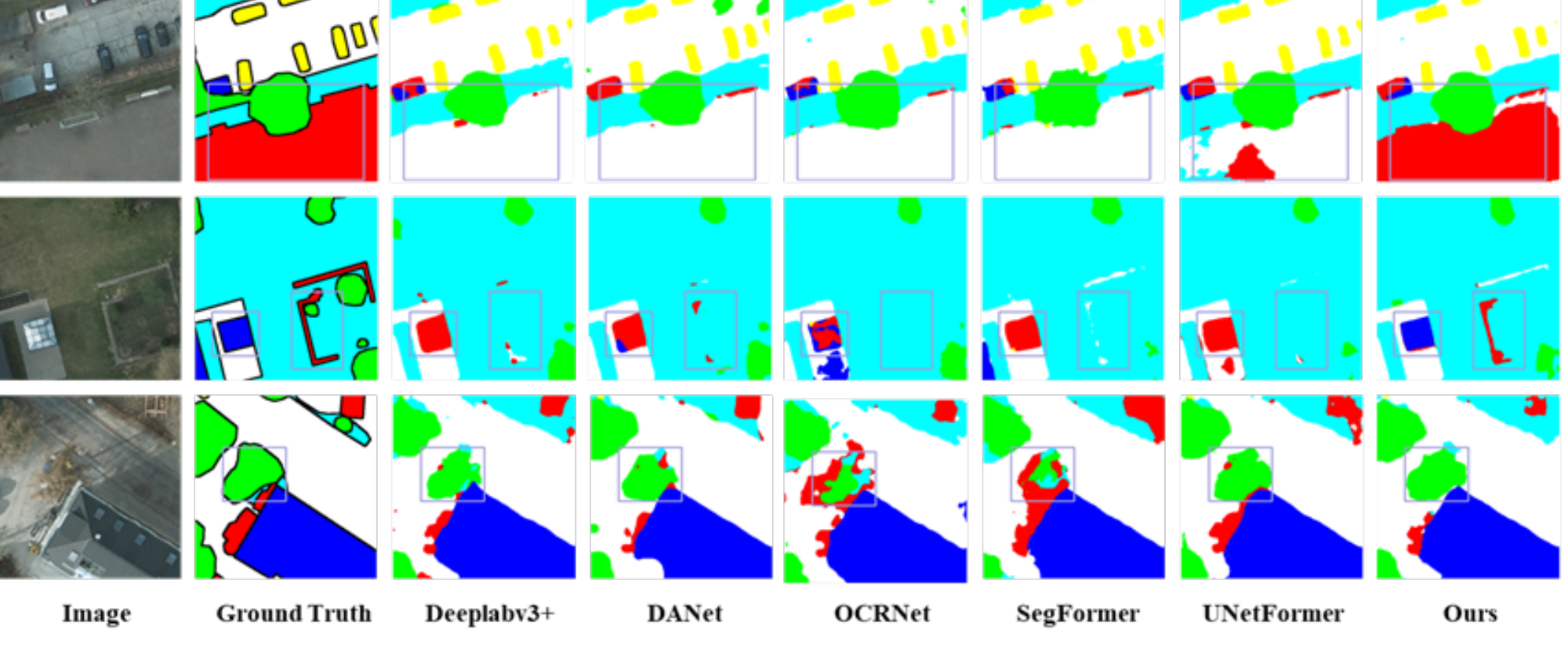}
	\caption{Visualization results on the potsdam dataset.}
	\label{fig_vis_pots}
\end{figure}

Specifically, our method achieves 75.48\% in mIoU and 81.23\% in mAcc for the Vaihingen dataset, and 77.90\% and 85.92\% for the Potsdam dataset, outperforming advanced CNN-based methods like DeeplabV3+. Benefiting from the category-based semantic information as well as spatial-detailed context, the performance of our method also surpasses recent Transformer-based approaches designed initially for remote sensing images, such as UNetFormer. Our AMMUNet achieved the best performance on three sub-categories in both datasets and obtained the highest overall scores. Visual examples can be found in Fig. \ref{fig_vis_vai} and Fig. \ref{fig_vis_pots}.

\subsection{Ablation Study}
To demonstrate the effectiveness of the proposed modules, we conducted ablation experiments on the Vaihingen dataset, comparing the performance of our approach with MSA, patch-free MSA, and GMSA+AMMM on both the U-Net\cite{ronneberger2015u} and UNetFormer architectures. The results, as shown in the Table \ref{tab:ablation_results}, indicate that our proposed method outperforms the others, achieving mIoU improvements of 0.39\% and 0.64\% on the U-Net and UNetFormer frameworks, respectively. 

\begin{table}[htbp]
	\centering
	\caption{ABLATION EXPERIMENTAL RESULTS OF VAIHINGEN DATASET}
	\resizebox{\linewidth}{!}{
		\begin{tabular}{lcccccc}
			\toprule
			Methods & Backbone & mIoU & mAcc & Param(M) & GFLOPs & FPS \\
			\midrule
			UNet & R50 & 72.68 & 80.09 & 63.45 & 116 & 72.46 \\
			UNet + MSA & R50 & 73.44 & 81.08 & 31.61 & 54.59 & 94.34 \\
			UNet + patch-free MSA & R50 & 73.53 & 80.99 & 47.77 & 83.57 & 30.86 \\
			UNet + GMSA + AMMM & R50 & 73.83 & 81.06 & 31.61 & 54.49 & 96.15 \\
			UNetFormer + MSA & R50 & 74.84 & 82.39 & 31.41 & 47.76 & 56.18 \\
			UNetFormer + patch-free MSA & R50 & 74.52 & 81.76 & 50.10 & 71.59 & 18.05 \\
			UNetFormer + GMSA + AMMM & R50 & 75.48 & 82.97 & 32.35 & 52.50 & 64.10 \\
			\bottomrule
		\end{tabular}
	}
	\label{tab:ablation_results}
\end{table}

Additionally, we evaluated the computational efficiency and inference speed of our method, comparing the model size, floating-point operations, and inference time with other approaches, as illustrated in the table. Our method maintains comparable inference speed and memory footprint to UNetFormer while reducing computational demands and inference time by approximately half compared to the patch-free MSA approach, maintaining similar parameter quantities and meeting real-time inference requirements. Our proposed method not only maintains accuracy advantages over this approach but also offers a smaller and more deployable parameter footprint.

\section{CONCLUSION}
In this letter, we propose AMMUNet, a novel U-Net-based architecture that employs ResNet as the encoder and integrates granular multi-head self-attention module with the attention map merging mechanism in the decoder. By avoiding window stacking, our model effectively balances the extraction and utilization of global and local information. Extensive experiments on the ISPRS Vaihingen and Potsdam datasets, benchmarked against existing methods, demonstrate the benefits of our class-guided mechanism and the superior segmentation accuracy achieved by the proposed approach.

\bibliographystyle{IEEEtran}
\bibliography{AMMUNet}

\begin{thebibliography}{10}
\providecommand{\url}[1]{#1}
\csname url@samestyle\endcsname
\providecommand{\newblock}{\relax}
\providecommand{\bibinfo}[2]{#2}
\providecommand{\BIBentrySTDinterwordspacing}{\spaceskip=0pt\relax}
\providecommand{\BIBentryALTinterwordstretchfactor}{4}
\providecommand{\BIBentryALTinterwordspacing}{\spaceskip=\fontdimen2\font plus
\BIBentryALTinterwordstretchfactor\fontdimen3\font minus
  \fontdimen4\font\relax}
\providecommand{\BIBforeignlanguage}[2]{{%
\expandafter\ifx\csname l@#1\endcsname\relax
\typeout{** WARNING: IEEEtran.bst: No hyphenation pattern has been}%
\typeout{** loaded for the language `#1'. Using the pattern for}%
\typeout{** the default language instead.}%
\else
\language=\csname l@#1\endcsname
\fi
#2}}
\providecommand{\BIBdecl}{\relax}
\BIBdecl

\bibitem{long2015fully}
J.~Long, E.~Shelhamer, and T.~Darrell, ``Fully convolutional networks for
  semantic segmentation,'' in \emph{Proc. IEEE Conf. Comput. Vis. Pattern
  Recognit. (CVPR)}, Boston, MA, USA, Jun. 2015, pp. 3431--3440.

\bibitem{he2016deep}
K.~He, X.~Zhang, S.~Ren, and J.~Sun, ``Deep residual learning for image
  recognition,'' in \emph{Proc. IEEE Conf. Comput. Vis. Pattern Recognit.
  (CVPR)}, Jun. 2016, pp. 770--778.

\bibitem{chen2017rethinking}
L.-C. Chen, G.~Papandreou, F.~Schroff, and H.~Adam, ``Rethinking atrous
  convolution for semantic image segmentation,'' \emph{arXiv preprint
  arXiv:1706.05587}, 2017.

\bibitem{chen2018encoder}
L.-C. Chen, Y.~Zhu, G.~Papandreou, F.~Schroff, and H.~Adam, ``Encoder--decoder
  with atrous separable convolution for semantic image segmentation,'' in
  \emph{Proc. Eur. Conf. Comput. Vis. (ECCV)}, Sep. 2018, pp. 801--818.

\bibitem{dosovitskiy2020image}
A.~Dosovitskiy \emph{et~al.}, ``An image is worth 16x16 words: Transformers for
  image recognition at scale,'' in \emph{International Conference on Learning
  Representations}, 2020.

\bibitem{vaswani2017attention}
A.~Vaswani \emph{et~al.}, ``Attention is all you need,'' in \emph{Proc. Conf.
  Neural Informat. Process. Syst. (NeurIPS)}, 2017, pp. 6000--6010.

\bibitem{liu2021swin}
Z.~Liu \emph{et~al.}, ``Swin transformer: Hierarchical vision transformer using
  shifted windows,'' in \emph{Proc. IEEE/CVF Int. Conf. Comput. Vis. (ICCV)},
  Oct. 2021, pp. 9992--10\,002.

\bibitem{li2021abcnet}
R.~Li, S.~Zheng, C.~Zhang, C.~Duan, L.~Wang, and P.~M. Atkinson, ``Abcnet:
  Attentive bilateral contextual network for efficient semantic segmentation of
  fine-resolution remotely sensed imagery,'' \emph{ISPRS J. Photogramm. Remote
  Sens.}, vol. 181, pp. 84--98, Nov. 2021.

\bibitem{li2022multiattention}
R.~Li \emph{et~al.}, ``Multiattention network for semantic segmentation of
  fineresolution remote sensing images,'' \emph{IEEE Trans. Geosci. Remote
  Sens.}, vol.~60, p. 21546232, 2022.

\bibitem{ronneberger2015u}
O.~Ronneberger, P.~Fischer, and T.~Brox, ``U-net: Convolutional networks for
  biomedical image segmentation,'' in \emph{Proc. Int. Conf. Med. Image Comput.
  Comput.-Assist. Intervent.}\hskip 1em plus 0.5em minus 0.4em\relax Cham,
  Switzerland: Springer, 2015, pp. 234--241.

\bibitem{cao2022swin}
H.~Cao \emph{et~al.}, ``Swin-unet: Unet-like pure transformer for medical image
  segmentation,'' in \emph{Computer Vision---ECCV 2022 Workshops, Tel Aviv,
  Israel}.\hskip 1em plus 0.5em minus 0.4em\relax Switzerland: Springer, Oct.
  2022, pp. 205--218.

\bibitem{he2022swin}
X.~He, Y.~Zhou, J.~Zhao, D.~Zhang, R.~Yao, and Y.~Xue, ``Swin transformer
  embedding unet for remote sensing image semantic segmentation,'' \emph{IEEE
  Trans. Geosci. Remote Sens.}, vol.~60, p. 4408715, 2022.

\bibitem{li2021multistage}
R.~Li, S.~Zheng, C.~Duan, J.~Su, and C.~Zhang, ``Multistage attention resu-net
  for semantic segmentation of fine-resolution remote sensing images,''
  \emph{IEEE Geoscience and Remote Sensing Letters}, pp. 1--5, 2021.

\bibitem{wang2022unetformer}
L.~Wang \emph{et~al.}, ``Unetformer: A unet-like transformer for efficient
  semantic segmentation of remote sensing urban scene imagery,'' \emph{ISPRS J.
  Photogramm. Remote Sens.}, vol. 190, pp. 196--214, Aug. 2022.

\bibitem{guo_attention_2022}
M.-H. Guo \emph{et~al.}, ``Attention mechanisms in computer vision: {A}
  survey,'' \emph{Computational Visual Media}, vol.~8, no.~3, pp. 331--368,
  2022.

\bibitem{su2023global}
J.-N. Su, M.~Gan, G.-Y. Chen, J.-L. Yin, and C.~L.~P. Chen, ``Global learnable
  attention for single image super-resolution,'' \emph{IEEE Trans. Pattern
  Anal. Mach. Intell.}, vol.~45, pp. 8453--8465, Jul. 2023.

\bibitem{wang_non-local_2018}
X.~Wang, R.~Girshick, A.~Gupta, and K.~He, ``Non-local {Neural} {Networks},''
  in \emph{2018 {IEEE}/{CVF} {Conference} on {Computer} {Vision} and {Pattern}
  {Recognition}}.\hskip 1em plus 0.5em minus 0.4em\relax Salt Lake City, UT,
  USA: IEEE, Jun. 2018, pp. 7794--7803.

\bibitem{guo_skysense_2023}
X.~Guo \emph{et~al.}, ``{SkySense}: {A} {Multi}-{Modal} {Remote} {Sensing}
  {Foundation} {Model} {Towards} {Universal} {Interpretation} for {Earth}
  {Observation} {Imagery},'' 2023.

\bibitem{fu2019dual}
J.~Fu \emph{et~al.}, ``Dual attention network for scene segmentation,'' in
  \emph{Proc. IEEE/CVF Conf. Comput. Vis. Pattern Recognit. (CVPR)}, Jun. 2019,
  pp. 3146--3154.

\bibitem{yuan2020object}
Y.~Yuan, X.~Chen, and J.~Wang, ``Object-contextual representations for semantic
  segmentation,'' in \emph{Computer Vision--ECCV 2020: 16th European
  Conference, Glasgow, UK, August 23--28, 2020, Proceedings, Part VI 16}, 2020,
  pp. 173--190.

\bibitem{xie2021segformer}
E.~Xie, W.~Wang, Z.~Yu, A.~Anandkumar, J.~M. Alvarez, and P.~Luo, ``Segformer:
  Simple and efficient design for semantic segmentation with transformers,'' in
  \emph{Advances in Neural Information Processing Systems}, vol.~34, 2021, pp.
  12\,077--12\,090.

\end{thebibliography}

\end{document}